# A Constraint-based Case Frame Lexicon

Kemal Oflazer and Okan Yılmaz
Department of Computer Engineering and Information Science
Bilkent University
Bilkent, Ankara 06533, Turkey
{ko,okan}@cs.bilkent.edu.tr

## Abstract

We present a constraint-based case frame lexicon architecture for bi-directional mapping between a syntactic case frame and a semantic frame. The lexicon uses a semantic sense as the basic unit and employs a multi-tiered constraint structure for the resolution of syntactic information into the appropriate senses and/or idiomatic usage. Valency changing transformations such as morphologically marked passivized or causativized forms are handled via lexical rules that manipulate case frames templates. The system has been implemented in a typed-feature system and applied to Turkish.

## 1 Introduction

Recent advances in theoretical and practical aspects of feature and constraint-based formalisms for representing linguistic information have fostered research on the use of such formalisms in the design and implementation of computational lexicons (Briscoe *et al.*, 1993). Case frame approach has been the representation of choice especially for languages with free constituent order, explicit case marking of noun phrases and embedded clauses filling nominal syntactic roles. The semantics of such syntactic role fillers are usually determined by their lexical, semantic and morphosyntactic properties, instead of position in the sentence. In this paper, we present an approach to building a constraint-based case frame lexicon for use in natural language processing in Turkish.

A number of observations that we have made on Turkish have indicated that we have to go beyond the traditional transitive and intransitive distinction, and utilize a framework where verb valence is considered as the obligatory co-existence of an arbitrary subset of possible arguments along with the obligatory exclusion of certain others, relative to a verb sense. Additional morphosyntactic, lexical and semantic selectional constraints are utilized to map a given syntactic argument structure to a specific verb sense. In recent years, there have been several studies on constraint-based lexicons. Russell *et al.* (1993) propose an approach to multiple default inheritance for unification-based lexicon. In another study by Lascarides *et al.* (1995), an ordered approach to default unification is suggested. de Paiva (1993) formalizes the system of well-formed typed feature structures. In this study, type hierarchies and relations are mathematically defined. They also formalize unification and generalization operators between the feature structures, along with defining well-formedness notion that we use in our system.

## 2 Representing Case Frame Information

In Turkish, (and possibly in many other languages) verbs often convey several meanings (some totally unrelated) when they are used with subjects, objects, oblique objects, adverbial adjuncts, with certain lexical, morphological, and semantic features, and co-occurrence restrictions. In addition to the usual sense variations due to selectional restrictions on verbal arguments, in most cases, the meaning conveyed by a case frame is idiomatic, with subtle constraints. For example, the Turkish verb *ye* (*eat*), when used with a direct object noun phrase whose head is:

1. *para* (*money*), with *no* case or possessive markings and a human subject, means *to accept bribe*,

2. *para* (*money*), with a non-human subject, means *to cost a lot*,

3. *para* (or any other NP whose head is ontologically IS-A money, e.g., *dolar, mark*, etc.) with *obligatory* accusative marking and *optional* possessive marking, means *to spend money*,

4. *kafa* (*head*) with *obligatory* accusative marking and *no* possessive marking, means *to get mentally deranged*,

5. *hak* (*right*) with optional accusative and possessive markings, means *to be unfair*,

6. *baş* (*head*, cf. 4) (or any NP whose head is ontologically IS-A human) with *optional* accusative and optional possessive marking (obligatory only with *baş*), means *to waste* or *demote a person*.

On the other hand:

1. if an *ablative case-marked oblique object* denoting an edible entity is present, then there should *not* be any direct object, and the verb means *to eat a piece of (the edible (oblique) object)*, or

2. if the ablative case-marked oblique object does not denote something edible, but rather a container, then the sense maps to *to eat out of*, with the *optional* direct (edible) object denoting the object eaten.

Clearly such usage has impact on thematic role assignments to various role fillers, and even on the syntactic behavior of the verb in question (Briscoe and Carroll, 1994). For instance, for the third and fourth cases above where the object has to be obligatorily case-marked accusative, a passive form would not be grammatical for the sense conveyed, although syntactically *ye (eat)* is a transitive verb.

Sometimes verbs require different combinations of arguments, or explicitly require that certain arguments not be present. For instance, the verb *şaş* requires different kinds of arguments depending on the sense, obligatorily excluding other arguments:

1. an *ablative case-marked oblique object* and with *no other* object in the case frame *şaş* means *to deviate from*,

2. a *dative case-marked oblique object* and with *no other object*, *şaş* means *to be surprised at*,

3. an *accusative case-marked* direct object with *no other object*, *şaş* means *to be confused about*.

As a final example, when the verb *tut* (catch/hold) is used with an *obligatory $3^{rd}$ person singular agreement* and *active voice*, and *the subject is a (nominalized) S with a verb form of future participle*, then the sense conveyed by the top level case frame is *to feel like doing* the predication indicated by the subject S's case frame, with the agent being the subject of this embedded clause.

As illustrated in these examples, verb sense idiomatic usage resolution has to be dealt with in a principled way and not by pattern matching (e.g., as in Tschichold (1995)), when the language has a free word order, where pattern matching approaches could fail. In this paper, we present a unification-based approach to a constraint-based case frame lexicon, in which one single mechanism deals with both problems uniformly. The essential function of our lexicon is to map bidirectionally between a case frame containing information that is syntactic, and a semantic frame which captures the predication denoted by the case frame along with information about who fills what thematic role in that predication.

## 3 The Lexicon Architecture

In this section we present an overview of structure of lexicon entries and the nature of the constraints. The basic unit in the lexicon is a *sense* which is the information denoting some *indivisible predication along with the thematic roles involved*. We generate the case frame of each sense by unifying a set of co-occurrence, morphological, syntactic, semantic, and lexical constraints on verbs, their arguments. The lexicon is implemented in TFS (Kuhn, 1993) by the disjunction of the senses defined by unifying `wf-case-frame` (well-formed case frame) with each sense:

```
wf-case-frame < case-frame.
wf-case-frame & SENSE#1.
wf-case-frame & SENSE#2.
...              ...
wf-case-frame & SENSE#n.
```

### 3.1 Lexicon Entries

Each verb sense entry in the lexicon has the structure shown by the feature structure matrix in Figure 1.

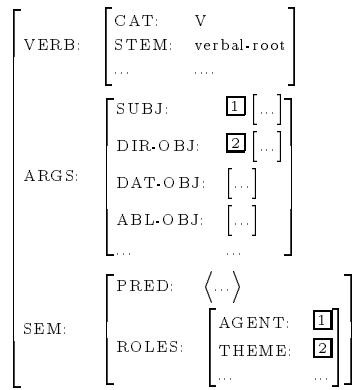

Figure 1: Structure of a case frame lexicon entry.

The feature structure for each syntactic argument contains information about the morphological and syntactic structure of the syntactic constituent such as part-of-speech, agreement, case, possessive markers, and additional morphological markings such as verb form, (e.g., infinitive, participle, etc.), voice (e.g., active, passive, causative, reflexive, reciprocal, etc.) for embedded S's, along with their own case frames. This structure is similar to the structure proposed in Lascarides *et al.* (1995). However, instead of classifying argument structures as simply transitive, intransitive, etc., we need to consider all relevant elements of the power set of possible arguments. For Turkish, the syntactic constituents that we have chosen to in-

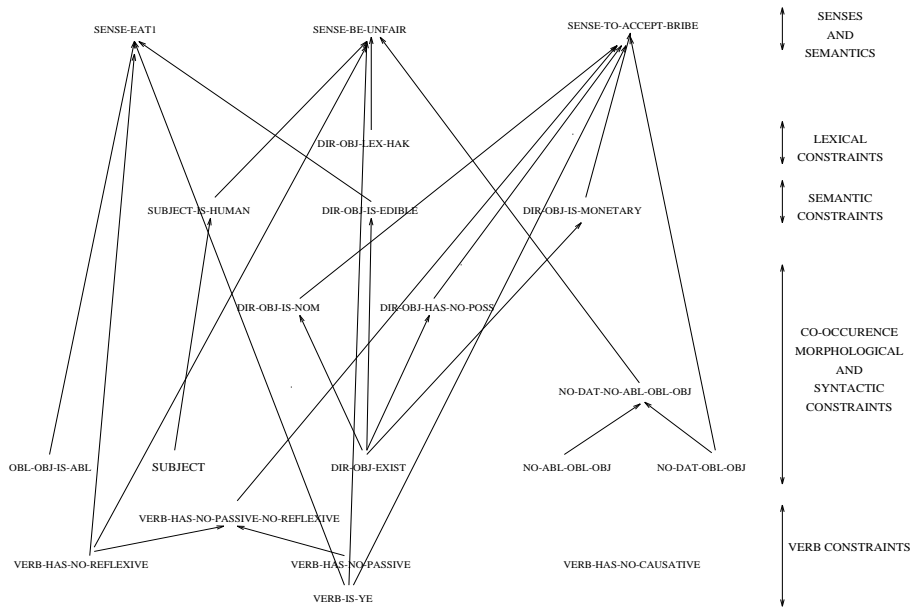

Figure 2: The portion of the constraint structure for a portion of the the Turkish verb *"ye"*.

clude in the argument slot (for a verb in active voice) are the following:

- subject (nominative `NP`),[1]
- direct object (nominative or accusative case-marked `NP`),
- oblique objects (ablative, dative, locative case-marked `NP`),
- beneficiary object (dative case-marked `NP`, or `PP` with a certain `PFORM`),
- instrument object (instrumental case-marked `NP` or `PP` with a certain `PFORM`),
- value object (dative case-marked `NP` or `PP` with a certain `PFORM`).

In general, there may be more than one instantiation of the `SEM` frame for a given instantiated set of case frame arguments (and vice versa). For instance, for the *ye* verb discussed above, the argument structure for the third case giving rise to the meaning *to get mentally deranged* may conceivably give rise to a literal meaning in a rather improbable context (such as eating the head of a fish at dinner - much in the spirit of the two interpretations of the English idiom *kick the bucket*), or the same semantics may be expressed by a different surface form.

### 3.2  Constraint Architecture

We express constraints on the arguments in the case frame of a verb via a 5-tier constraint hierarchy sharing constraints among the specification of other constraints and sense definitions, whenever possible:

---

[1] NP's that have no case-marking in Turkish.

1. *Constraints on verb features* that describe any relevant constraints on the morphological features of the verb, such as agreement or voice markers.

2. *Constraints on morphological features* that describe any *obligatory* constraints on the arguments, such as case-marking, verb form (in the case of embedded clauses), etc.

3. *Constraints on argument co-occurrence* that express *obligatory* argument co-occurrence constraints along with constraints that indicate when certain arguments should *not* occur in order resolve a sense.

4. *Lexical constraints* that indicate any specific constraints on the heads of the arguments in order to convey a certain sense, and usually constrain the stem of the head noun to be a certain lexical form, or one of a small set of lexical forms.

5. *Semantic Constraints* that indicate semantic selectional restriction constraints that may resolved using a companion ontological database (again implemented in TFS) in which we model the world by defining semantic categories, such as *human*, *thing*, *non-living object*, *living object*, etc., along the lines described by Nagao *et al.* (1985).

Figure 2 illustrates the simplified form of the constraint–sense mapping of the verb *ye (eat)*.

### 3.3  Valency Changing Transformations

As we have already stated, we encode senses of verbs in active voice unless a verb has an idiomatic usage with obligatory passive, causative and/or reflexive voices.[2] In order to handle these valency changing transformations, we define lexical rules as shown in Figure 3.

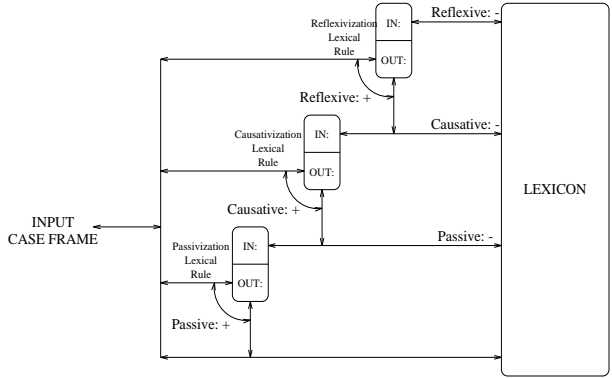

Figure 3: Valency transformations using lexical rules.

This figure describes how a given case frame with its syntactic constituents is processed by a sequence of lexical rules each stripping off a certain voice marker and then attempting unification with the lexicon for any possible sense resolution. The order of lexical rules in this figure reflects the reverse order of voice markers in Turkish verbal morphology.[3] So a given case frame may have to go through three lexical rules until it finds a unifying entry in the lexicon. Unifications before going through all lexical rules are for (possibly idiomatic) senses which explicitly require various voice markings. Two additional constituents are added via these lexical rules. The `AGN-OBJ` (agentive object), denotes the equivalent of the *by-object* in passive sentences. The subject of the sentences a causative voice marked verb is indicated by `CAUSER` in the semantics frame. Our current implementation does not deal with multiple causative voice markings (which Turkish allows), or with the rather tricky surface case change of the object of causation depending on the transitivity of the causativized verb. In the examples and sample rules below, a voice marker can take one of three values: (i) `+`: indicates the voice marker has to be taken. (ii) `-`: indicates the voice marker is not taken (iii) `nil`: indicates the voice marker must not be taken; this is used only in the sense definitions in the lexicon and can unify with `-` but not with `+`.

---

[2]For instance:
| birine | vurmak | vs. |
| someone+DAT | hit+INF | |
| to hit | someone | |

| someone+DAT | hit+PASS+INF | |
| birine | vurulmak | |
| to fall in love with | someone | |

[3]We have not dealt with the reciprocal/collective voice marker yet.

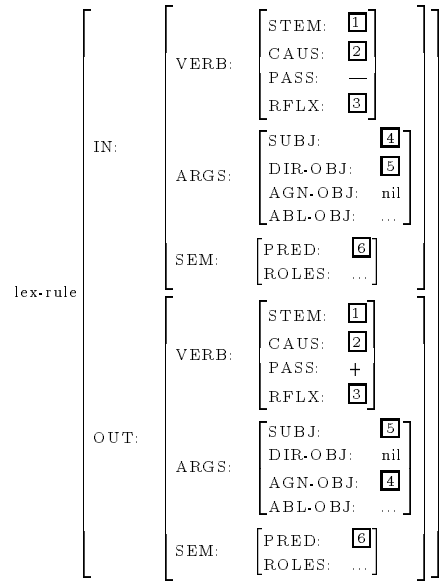

Figure 4: The simplified passivization rule for transitive verbs

Figures 4 and 5 show two of the simpler lexical rules.

### 3.4 Examples

In this section we present a few examples that show how one can describe a given verb sense. For the first example the following constraints are employed:

1. `VERB-IS-YE` is a constraint corresponding to [VERB: | STEM: "ye"]

2. `VERB-TAKES-NO-PASSIVE-NO-REFLEXIVE` is the verb constraint [VERB: [PASSIVE: nil, RFLX: nil]]

3. `DIR-OBJ-HAS-NO-POSS` is the morphological constraint [ARGS: | DIR-OBJ: | POSS: none]

4. `DIR-OBJ-IS-ACC` is the morphological constraint [ARGS: |DIR-OBJ: |CASE: acc]

5. `NO-DATIVE-OBL-OBJ` is the argument co-occurrence constraint [ARGS: |DAT-OBL: nil]

6. `SUBJECT-IS-HUMAN` is the semantic constraint [ARGS: |SUBJECT: |HEAD: |SEM: human]

7. `DIR-OBJ-HEAD-LEX-KAFA` is a lexical constraint [ARGS: |DIR-OBJ: |HEAD: |LEX: "kafa"]

8. `SEM-GET-MENTALLY-DERANGED` is the feature structure for the semantics portion

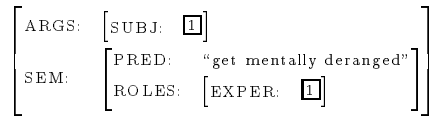

We can then express the constraint for the verb sense by unifying (denoted by `&` in TFS) all the

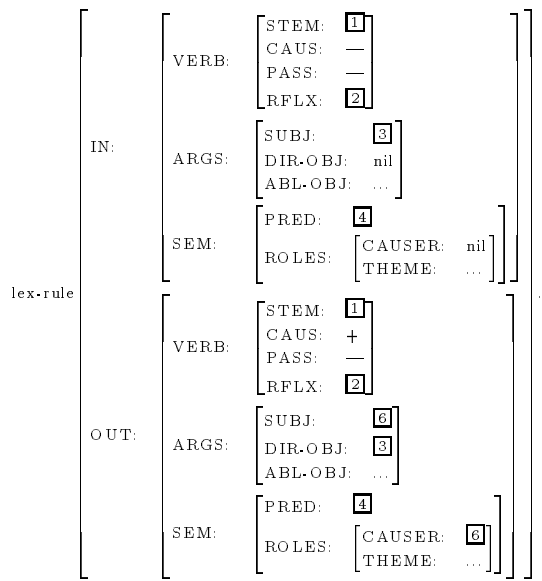

Figure 5: The simplified causation rule for intransitive verbs

constraints above:

```
SENSE-GET-MENTALLY-DERANGED :=
  VERB-IS-YE &
  VERB-TAKES-NO-PASSIVE-NO-REFLEXIVE &
  DIR-OBJ-HAS-NO-POSS & DIR-OBJ-IS-ACC &
  NO-DATIVE-OBL-OBJ & DIR-OBJ-LEX-KAFA &
  SUBJECT-IS-HUMAN &
  SEM-GET-MENTALLY-DERANGED.
```

The resulting constraint when unified with partially specified case frame entry – an entry where only the argument and verb entries have been specified, will supply the unspecified SEM component(s). That is, when a partially specified case frame such as unifies successfully with the given constraint above, the unspecified portion will be properly instantiated with the experiencer being coindexed with the subject in the arguments.

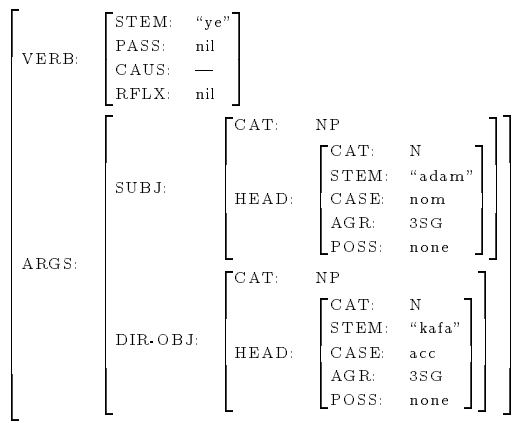

As a second example, consider the default sense of *ye* corresponding *to eat (something)*. The constraints are:

1. `VERB-IS-YE` is the verb constraint
   $[\text{VERB: | STEM: "ye"}]$

2. `VERB-TAKES-NO-REFLEXIVE` is the verb constraint $[\text{VERB: | RFLX: nil}]$

3. `NO-DAT-OBL-OBJ` is the co-occurrence constraint $[\text{ARGS: |DAT-OBL: nil}]$

4. `DIR-OBJ-IS(optional-edible)` is the disjunctive argument constraint

   $\left[\text{ARGS: |DIR-OBJ: }\left\{\begin{matrix}[\text{HEAD: }[\text{SEM: edible}]]\\ \text{nil}\end{matrix}\right\}\right]$

   (This is just explanatory, see below for how this is implemented in TFS.)

5. `ABL-OBJ-IS(optional-container)` is the argument constraint

   $\left[\text{ARGS: |ABL-OBJ: }\left\{\begin{matrix}[\text{HEAD: }[\text{SEM: container}]]\\ \text{nil}\end{matrix}\right\}\right]$

6. `INST-OBJ-IS(optional-instrument)` is the argument constraint

   $\left[\text{ARGS: |INST: }\left\{\begin{matrix}[\text{HEAD: }[\text{SEM: instrument}]]\\ \text{nil}\end{matrix}\right\}\right]$

7. `SEM-EAT1` is the feature structure for the semantics portion

   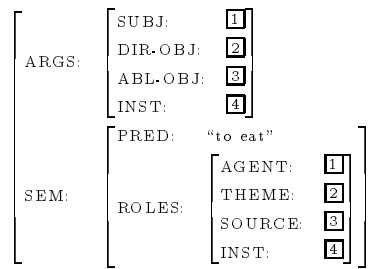

In most cases, there are arguments that are not obligatorily required for resolving a verb sense. These, nevertheless, have to be constrained, usually on semantic grounds. For instance the direct object is not obligatory for the basic sense of *ye*, but has to be an edible entity if it is present. We handle these constraints by defining a slightly more complex type hierarchy:

```
argument = noun-phrase |
           case-frame |
           optional.
optional = optional-edible |
           optional-container |
           optional-instrument. ...
optional-edible = nil | edible-obj.
edible-obj &   noun-phrase & IS-A-EDIBLE.
```

where `IS-A-EDIBLE` is a constraint of the form $[\text{HEAD: | SEM: edible}]$. The optional ablative and instrumental objects are defined similarly.[4] The

---
[4]Note that the surface case constraints for these are defined in the basic definition of the case frame.

sense definition then becomes:

```
SENSE-EAT1 :=
  VERB-IS-YE &  VERB-TAKES-NO-REFLEXIVE &
  NO-DATIVE-OBL-OBJ &
  DIR-OBJ-IS(optional-edible) &
  ABL-OBL-OBJ(optional-container) &
  INST-OBJ-IS(optional-instrument) & SEM-EAT1.
```

As a more complicated example employing nested clauses, we present below the case frame for the last example in Section 2, where the verb *tut* (*catch*) is used with a clausal subject for a very specific idiomatic usage.

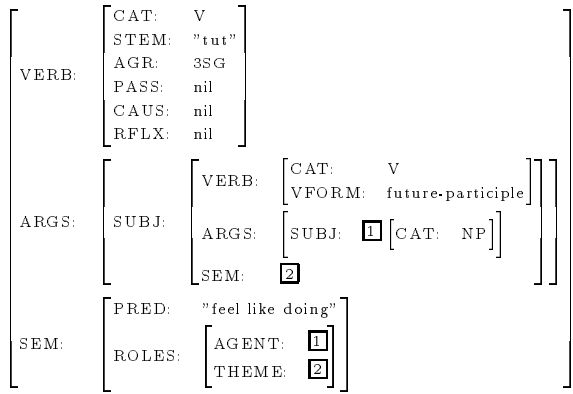

In this case, *the sense resolution of the embedded case frame is also performed concurrently with the case frame resolution of the top-level frame.*

The last example below illustrates the handling of valency changing transformations where lexical rules handle argument shuffling.

| Çocuk | adam | tarafından |
|---|---|---|
| *Child* | *man* | *by* |

| karşıya | geçirildi. | |
|---|---|---|
| *opposite_side* | *pass*+CAUS | |
| +DAT | +PASS+PAST+3SG | |

(The child was passed to the opposite side by the man.)

The output for this sentence is presented on the right.

## 4 Conclusions

This paper has presented a constraint-based lexicon architecture for representing and resolving verb senses and idiomatic usage in a case frame framework using constraints on different dimensions of the information available. Economy of representation is achieved via sharing of constraints across many verb sense definitions. The system has been implemented using the TFS system.

## 5 Acknowledgments

This research was in part funded by a NATO Science for Stability Phase III Project Grant – TU-LANGUAGE.

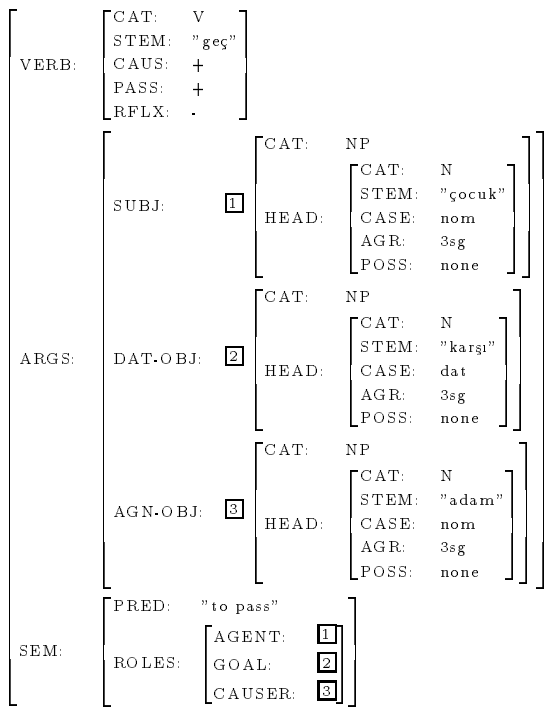

## References

Briscoe, E.J., A. Copestake, and V. de Paiva (eds.). 1993. *Inheritance Defaults and the Lexicon*. Cambridge University Press.

Briscoe, E. J. and J. Carroll 1994. *Towards Automatic Extraction of Argument Structure from Corpora* Technical Report, MLTT-006, Rank Xerox Research Centre, Grenoble.

Kuhn, J. 1993. *Encoding HPSG Grammars in TFS*. Institut für Maschinelle Sprachverarbeitung, Universität Stuttgart, Germany, March.

Lascarides, A., T. Briscoe, N. Asher, and A. Copestake. 1995. Order Independent and Persistent Typed Default Unification, Technical Report, Cambridge University, Computer Laboratory, March.

Nagao, M., J. Tsujii, and J. Nakamura. 1985. The Japanese Government Project for Machine Translation. In *Computational Linguistics*, volume 11. April-September.

de Paiva, V. 1993. Types and Constraints in LKB. In Briscoe *et al.* (1993).

Russell, G , A. Ballim, J. Carroll, and S. Warwick-Armstrong. 1993. A Practical Approach to Multiple Default Inheritance for Unification-Based Lexicons. In In Briscoe *et al.* (1993).

Tschichold, C. 1995. English Multi-word Lexemes In A Lexical Database. In *Proceedings of the Lexicon workshop of ESSLLI'95, Seventh European Summer School in Logic Language and Information*, August